\documentclass[11pt]{article}

\usepackage[final]{acl}

\usepackage{times}
\usepackage{latexsym}
\usepackage{amsmath}
\usepackage{amssymb}
\usepackage{multirow} 
\usepackage[T1]{fontenc}

\usepackage[utf8]{inputenc}

\usepackage{microtype}

\usepackage{inconsolata}

\usepackage{graphicx}

\title{Extensible Multi-Granularity Fusion Network and Transferable Curriculum Learning for Aspect-based Sentiment Analysis}

\author{
  \textbf{Xinran Li\textsuperscript{1,2}},
  \textbf{Xiaowei Zhao\textsuperscript{1}},
  \textbf{Yubo Zhu\textsuperscript{1}},
  \textbf{Zhiheng Zhang\textsuperscript{1}}, \\
  \textbf{Zhiqi Huang\textsuperscript{2}},
  \textbf{Hongkun Song\textsuperscript{2}},
  \textbf{Jinglu Hu\textsuperscript{2}},
  \textbf{Xinze Che\textsuperscript{1}}, 
  \textbf{Yifan Lyu\textsuperscript{1}},\\
  \textbf{Yong Zhou\textsuperscript{1}\thanks{Corresponding author.}},
  \textbf{Xiujuan Xu\textsuperscript{1}\footnotemark[1]}
\\
\\
  \textsuperscript{1}School of Software Technology, Dalian University of Technology \\
  \textsuperscript{2}Graduate School of Information, Production and Systems, Waseda University
    \\
  {\small \textbf{Correspondence:} 
  \href{mailto:tyzy8999@gmail.com}{tyzy8999@gmail.com},
  \href{mailto:xjxu@dlut.edu.cn}{xjxu@dlut.edu.cn}}
}

\begin{document}
\maketitle
\begin{abstract}
Aspect-based Sentiment Analysis (ABSA) aims to determine sentiment polarity toward specific aspects in text. Existing methods enrich semantic and syntactic representations through external knowledge or GNNs, but the growing diversity of linguistic features increases model complexity and lacks a unified, extensible framework. We propose an Extensible Multi-Granularity Fusion Network (EMGF) that integrates dependency syntax, constituent syntax, attention-based semantics, and external knowledge graphs. EMGF employs multi-anchor triplet learning and orthogonal projection to effectively fuse multi-granularity features and strengthen their interactions without additional computational overhead. Furthermore, we introduce the first task-specific curriculum learning framework for text-only ABSA, which assigns difficulty scores using five indicators and trains the model from easy to hard to mimic human learning and improve generalization. Experiments on SemEval 2014, Twitter, and MAMS datasets show that EMGF+CL consistently outperforms state-of-the-art ABSA models.
\end{abstract}

\section{Introduction}

The primary goal of Aspect-Based Sentiment Analysis (ABSA) is to identify the sentiment polarity toward specific aspects or entities in a text, enabling a finer-grained understanding of sentiment information. For example, in the review "\emph{\textbf{Looks} nice, but has a horribly cheap \textbf{feel} .}", the aspects "\emph{Looks}" and "\emph{feel}" express positive and negative sentiments, respectively. Unlike sentence-level sentiment analysis that assigns a single sentiment label to an entire sentence, ABSA can accurately capture the sentiment orientation of individual aspects. The main challenge is modeling aspect–opinion relationships effectively.

\begin{figure}[t]
    \centering
    \includegraphics[width=1.0\columnwidth]{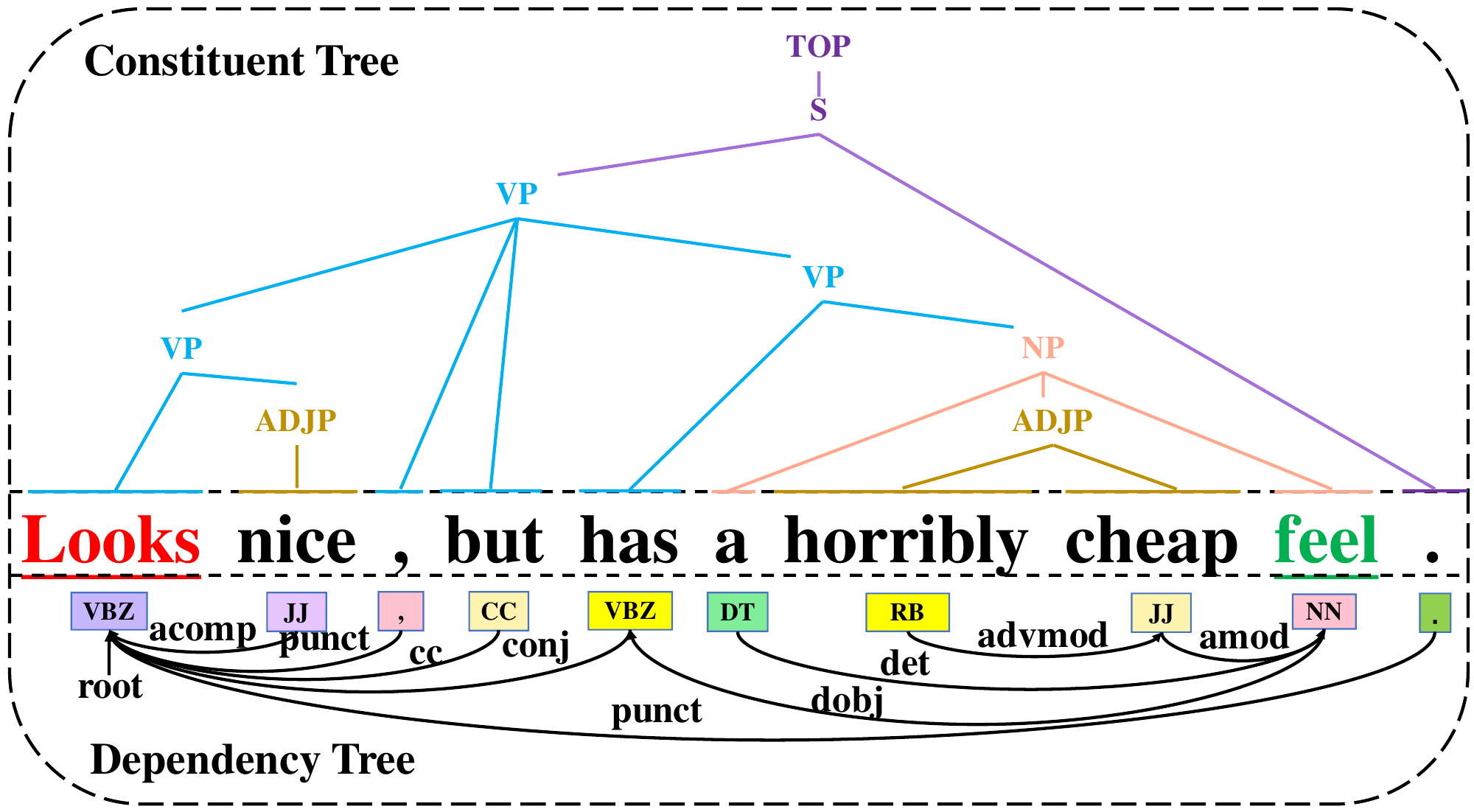}
    \caption{
    The sentence from a laptop review shows its dependency and constituent trees, containing two aspects with opposite sentiment polarities.
    }
    \label{Con_and_Dep}
\end{figure}

To enhance semantic representations in ABSA, previous studies have incorporated external knowledge and syntactic information \cite{DBLP:conf/aaai/MaPC18, DBLP:journals/kbs/ZhouHHH20, DBLP:journals/tkde/ZhongDLDJT23, zhang-etal-2019-aspect, sun-etal-2019-aspect, chen-etal-2020-inducing, liang-etal-2020-jointly, wang-etal-2020-relational, li-etal-2021-dual-graph, liang-etal-2022-bisyn}. For example, \citet{DBLP:journals/kbs/ZhouHHH20} enhanced semantic representations by constructing subgraphs with knowledge-graph-related words as seed nodes, though this approach becomes computationally expensive when there are many aspect terms; \citet{DBLP:journals/tkde/ZhongDLDJT23} embedded knowledge graphs into low-dimensional spaces to efficiently capture aspect-specific information. In addition, Graph Neural Networks (GNNs) have been applied to dependency (Dep.) and constituent (Con.) trees, modeling word-level relations and phrase-level hierarchies, respectively, which help align aspect terms with opinion words (as shown in Figure~\ref{Con_and_Dep}, the dependency link between “\emph{Looks}” and “\emph{nice}” and the phrase boundary “\emph{but}”). Although single-granularity features are effective in ABSA, they cannot fully capture the richness of textual information; \citet{li-etal-2021-dual-graph} demonstrated the benefits of multi-granularity fusion by integrating SynGCN and SemGCN through a Mutual BiAffine module. However, existing approaches remain complex and inefficient, lacking a scalable framework to unify syntactic, semantic, and external knowledge features. This raises a key question: how can multi-granularity information achieve cumulative performance gains while ensuring model scalability?

Although multi-granularity fusion enriches representations, optimizing the training process in ABSA remains underexplored. Curriculum learning \cite{CL} has been widely applied in various NLP tasks \cite{LSDGNN}, but its use in ABSA is limited. Existing methods often rely on complex models or multi-module fusion, which increase computational cost but yield only marginal gains. Moreover, ABSA datasets contain samples of varying difficulty, yet training typically treats all instances equally, making it hard to capture inherent differences effectively. Therefore, introducing a curriculum learning paradigm tailored for ABSA is a promising approach to enhance model robustness and generalization.

In this paper, we propose a novel architecture, the Extensible Multi-Granularity Fusion Network (EMGF), along with a transferable curriculum learning framework to address the aforementioned challenges. We enhance ABSA representation learning by integrating dependency syntax, constituent syntax, semantic attention, and external knowledge graphs. Meanwhile, we design an Extensible Multi-Stage Fusion (EMSF) module to capture complex multi-granularity interactions with low computational cost. The module consists of two stages: in the preprocessing stage, multi-anchor triplet learning is employed to jointly optimize dependency and constituent features, and orthogonal projection is used to enhance discriminability; in the fusion stage, external knowledge graphs further enrich the overall representation. During model training, we introduce a task-specific curriculum learning framework and innovatively design five difficulty indicators tailored for ABSA to rank the samples, training the model from easy to hard, thereby improving training effectiveness and model generalization. Our main contributions are summarized as follows:

\textbf{1)} We propose an Extensible Multi-Granularity Fusion Network (EMGF) for ABSA, which effectively captures cross-granularity interactions with low computational cost.

\textbf{2)} We introduce a task-specific curriculum learning framework for ABSA, in which five difficulty indicators are designed to rank samples from easy to hard.

\textbf{3)} Experiments on a range of datasets demonstrate that EMGF+CL achieves SOTA performance. The code will be publicly released.

\section{Related Work}

\begin{figure*} [t]
	\centering
	\includegraphics[width=2.0\columnwidth]{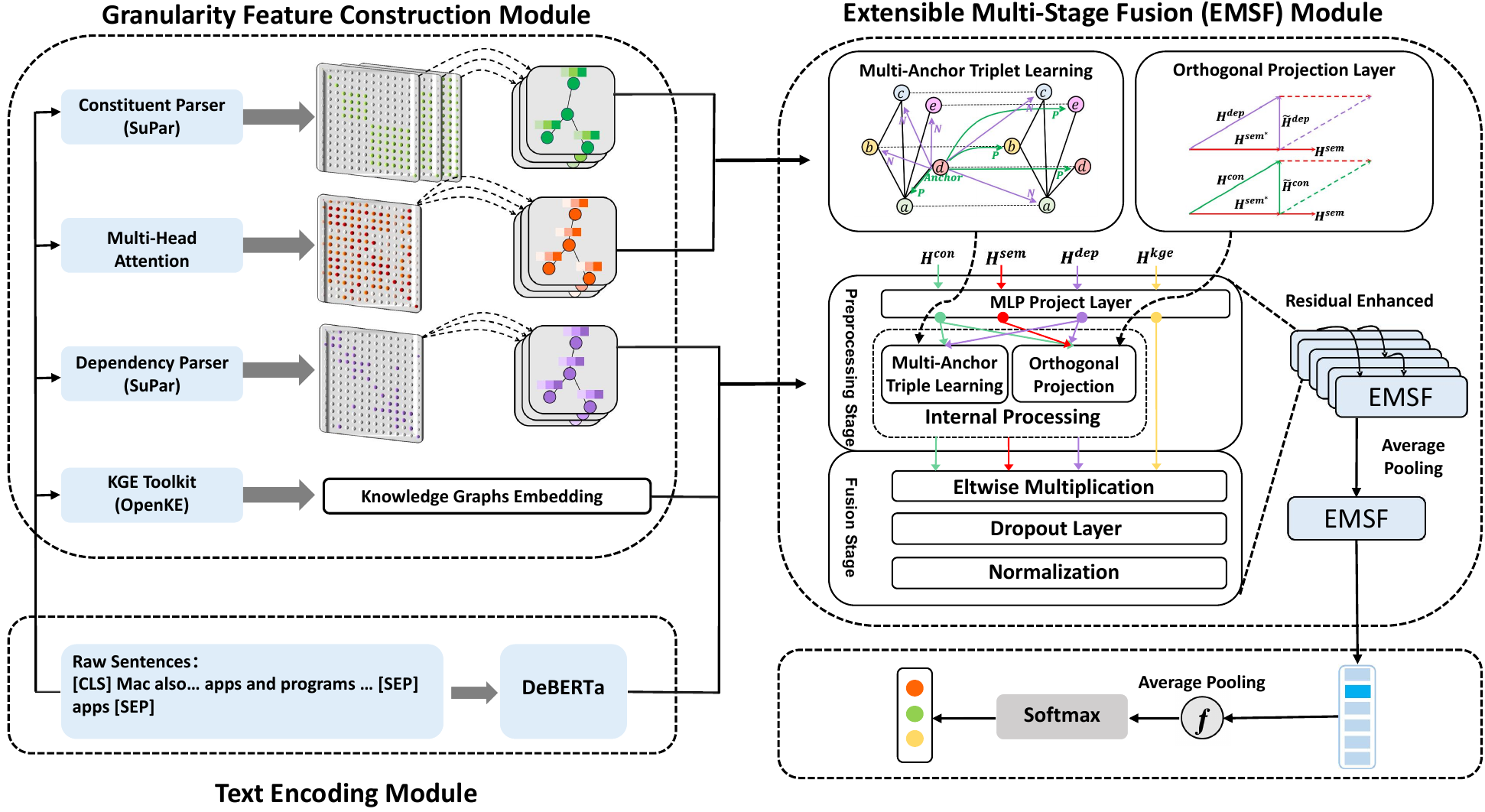}
	\caption{The overall architecture of our EMGF model.}
	\label{framework}
\end{figure*}

Recent methods mainly focus on two directions: traditional methods and LLM-based methods.

\textbf{Traditional Methods:} Recent approaches integrate syntactic information with GNNs, including dependency-based and constituent-based structures. Dependency-based methods use heterogeneous GCNs or dependency-augmented frameworks to capture aspect–opinion relations, sometimes enhanced with syntactic masks or refined importance scores \cite{DBLP:journals/access/XuZL20, DBLP:journals/ijon/LiangMZCXZ21, zhang-etal-2022-ssegcn, DBLP:journals/corr/abs-2311-04467}. Constituent-based methods employ GCNs to encode phrase structures and jointly learn syntactic-semantic relations, effectively modeling aspect-level sentiment context \cite{marcheggiani-titov-2020-graph, fei-etal-2021-better, liang-etal-2022-bisyn}.

\textbf{LLM-based Methods:} 
With the advent of InstructABSA \cite{InstructABSA}, ABSA has entered the era of large language models. In related work, CausalEdit-ABSC \cite{Li_Wang_Zhao_Zhang_Li_2025} leverages model editing to update only the intermediate representations critical for aspect-based sentiment classification, achieving efficient and interpretable fine-tuning; meanwhile, CD-ABSA \cite{LLaMACD} progressively removes word-level noise—including lexicographic, bag-of-words, and syntactic noise—to effectively enhance the prediction of sentiment polarity for specific aspect terms.

\textbf{Curriculum Learning:} 
\citet{LANGE2026129669} firstly explore curriculum learning within the ABSA task. Their method derives sample difficulty from SentiWordNet–based lexical scores and trains the model using baby-steps and one-pass curricula. However, their difficulty estimation relies on generic sentence-level sentiment features and an auxiliary classifier, without designing ABSA-specific difficulty indicators.

\section{Methodology}

This section details the EMGF framework illustrated in Figure \ref{framework}, comprising four parts: (1) Text Encoding, (2) Granularity Feature Construction, (3) Extensible Multi-Stage Fusion, and (4) Curriculum Learning.

\subsection{Text Encoding Module}  
In the ABSA task, given an $n$-word sentence $s=\{w_{1}, w_{2}, \dots, w_{n}\}$ and a specific aspect $a=\{a_{1}, a_{2}, \dots, a_{m}\}$, the goal is to predict its sentiment polarity class $c_{a}$, where $a$ is a sub-sequence of $s$, and $c_{a} \in \{Positive, Neutral, Negative\}$. To obtain contextualized embeddings, we employ \textbf{DeBERTaV3} \cite{he2023debertav}, which improves BERT through disentangled attention and ELECTRA-style pre-training. In this encoder, the sentence–aspect pair is constructed as $x = ([\mathrm{CLS}], s, [\mathrm{SEP}], a, [\mathrm{SEP}])$. The encoder outputs contextual embeddings $H^{\text{deberta}} = \mathrm{DeBERTaV3}(x)$, where $H^{\text{deberta}} = [h_{1}^{\text{deberta}}, h_{2}^{\text{deberta}}, \dots, h_{n}^{\text{deberta}}] \in \mathbb{R}^{n \times d}$, $d$ denotes the hidden size of the last layer, and $h_{i}^{\text{deberta}}$ represents the contextual embedding of the $i$-th word.

\subsection{Granularity Feature Construction Module}

\paragraph{Dependency GCN} The dependency graph convolutional networks (DepGCN) module takes syntactic encoding as input and utilizes the probability matrix of all dependency arcs from a dependency parser to encode syntax information. The dependency graph is embodied as an adjacency matrix $A^{\text{dep}} \in \mathbb{R}^{n \times n}$, which is defined as follows:
\begin{equation}
	\begin{split}
		A_{ij}^{\text{dep}}=\left\{\begin{array}{ll}
        1, & \text { if link}(i, j)=1 \\
        0, & \text { otherwise }
        \end{array}\right.
	\end{split}
\end{equation}
where $\text{link}(i, j)$ shows that $i$-th and $j$-th words have a dependence link. The dependency graph representation $H^{\text{dep}}=\{h_{1}^{\text{dep}}, h_{2}^{\text{dep}}, \ldots, h_{n}^{\text{dep}}\}$ is then obtained from the DepGCN module using the following formula:
\begin{equation}
	\begin{split}
		h_{i}^{l}=\sigma\Big(\sum_{j=1}^{n} A_{i j} W^{l} h_{j}^{l-1}+b^{l}\Big)
	\end{split}
        \label{eq:gcn_update}
\end{equation}
here, $W^{l}$ represents a weight matrix, $b^{l}$ denotes a bias term, and $\sigma$ is an activation function, such as ReLU.

\paragraph{Constituent GCN} \label{Con-GCN} We follow the syntax structure of the Con.Tree in a bottom-up manner, inspired by BiSyn-GAT+ \cite{liang-etal-2022-bisyn}. The Con.Tree is composed of multiple phrases (${ph}_{u}^{l}$) that make up the input text, and we create corresponding graphs based on these phrases ${ph}_{u}^{m}$.

Given the substantial depth of the constituent tree, we choose a total of $m$ layers with alternating intervals; for example, we select layer 1, skip layer 2, pick layer 3, and continue this pattern. We make this choice because the variation in phrase hierarchical information between adjacent layers is minimal, and excessive alignment would be an inefficient use of computational resources. Additionally, the chosen value of $m$ aligns with the number of ConGCN layers.

The constituent graph is embodied as an adjacency matrix $A^{\text{con}} \in \mathbb{R}^{l_{c} \times n \times n}$, which is defined as follows:
\begin{equation}
	\begin{split}
	\mathbf{A}_{i, j}^{\text{con(m)}}=\left\{\begin{array}{ll}1 & \text{if}\ w_{i}\  \text{and}\  w_{j} \text { are in same                   ${ph}_{u}^{m}$,} \\ 0 & \text { otherwise }\end{array}\right.
	\end{split}
\end{equation}

where $m$ denotes the level of the phrase within the selected $l_{c}$ layers, while $u$ denotes the constituent label associated with the phrase, such as S, NP, VP, and so on. Subsequently yields the output hidden representation $H^{\text{con}}=\left\{h_{1}^{\text{con}}, h_{2}^{\text{con}}, \ldots, h_{n}^{\text{con}}\right\}$ is then obtained from the ConGCN module using Eq. (\ref{eq:gcn_update}).

\paragraph{Semantic GCN} To construct the attention score matrix $A^{\text{sem}}$, we employ the Multi-Head Attention (MHA) mechanism on the hidden state features $H^{\text{deberta}}$ derived from the DeBERTa encoder. The MHA computes attention scores among words, and the formulation of the attention score matrix $A^{\text{sem}} \in \mathbb{R}^{n \times n}$ is as follows:
\begin{equation}
    A^{\text{sem}}_{ij} = Softmax(\text{MHA}(h_{i}^{\text{deberta}}, h_{j}^{\text{deberta}}))
\end{equation}
Subsequently yields the output hidden representation $H^{\text{sem}}=\left\{h_{1}^{\text{sem}}, h_{2}^{\text{sem}}, \ldots, h_{n}^{\text{sem}}\right\}$ is then obtained from the SemGCN module using Eq. (\ref{eq:gcn_update}).

\paragraph{External Knowledge} \citet{DBLP:journals/tkde/ZhongDLDJT23} synergistically combine contextual and knowledge information to achieve more comprehensive feature representations. We introduce the external knowledge as proposed by them, represented as $H^{\text{kge}}=\{h_{1}^{\text{kge}}, h_{2}^{\text{kge}}, \ldots, h_{n}^{\text{kge}}\}.$

\subsection{Extensible Multi-Stage Fusion Module}
Previous studies typically integrate only two granular features, making existing models unsuitable when additional features are incorporated. Building a model that supports multiple granularities simultaneously would substantially increase complexity and computational cost.

To address this limitation, we propose the Extensible Multi-Granularity Fusion (EMGF) module, which efficiently integrates diverse granular information while capturing complex cross-granularity interactions. EMGF is constructed by cascading multiple Extensible Multi-Stage Fusion (EMSF) blocks, each consisting of a preprocessing stage and a fusion stage. During the preprocessing stage, four types of features from different granular levels are used as inputs: $H^{\text{con}}$, $H^{\text{dep}}$, $H^{\text{sem}}$, and $H^{\text{kge}}$.

\subsubsection{Preprocessing Stage} 

Con.Tree and Dep.Tree share syntactic information from different viewpoints \cite{dong-etal-2022-syntactic}. \cite{DBLP:journals/tkdd/AtaFWSKL21, dong-etal-2022-syntactic} use multi-view learning to study three relationship categories: intra-node intra-view, intra-node inter-view, and inter-node inter-view. We collectively label nodes in these scenarios as "important nodes." However, there is currently no research addressing how to handle "non-important nodes," which could potentially disrupt the complementary learning of "important nodes." Moreover, to handle these three types of collaboration, it's necessary to design three distinct loss functions, adding complexity to the model. To this end, we propose Multi-Anchor Triplet Learning to address the two categories of issues mentioned above.

Inspired by \citet{qin-etal-2020-feature}, we apply orthogonal projection to enforce DepGCN and ConGCN to extract distinct syntactic information from the semantic representations produced by SemGCN, yielding more refined and discriminative syntactic–semantic features. In this stage, we integrate Multi-Anchor Triplet Learning with the orthogonal projection technique to capture complementary and discriminative properties across multiple granularity levels.

\paragraph{Multi-Anchor Triplet Learning} We choose a node from the con-view graph as the Anchor node and define the following as "pos" samples: \textbf{1)} nodes connected to the Anchor in the con-view; \textbf{2)} nodes in the dep-view that are homologous (i.e., correspond to the same entity in different views) to the Anchor; \textbf{3)} homologous dep-view counterparts of all con-view "pos" nodes, even when they are not linked to the Anchor’s homologous node. All remaining nodes are treated as "neg". The same procedure applies when the Anchor is chosen in the dep-view.

It is important to note that graph nodes vary in significance, and treating all nodes as Anchor nodes would weaken discrimination and precision. Following MP-GCN \cite{DBLP:journals/access/ZhaoXW22}, we adopt a Multi-Head S-Pool strategy to select Anchor nodes. Using the attention matrix $A^{\text{sem}}$, we perform both average and max pooling from two perspectives and select the Top-K nodes with the highest scores.

Our objective is to pull the Anchor node closer to the “pos” nodes while pushing it away from the “neg” nodes, which is achieved by minimizing:
\begin{equation}
        \begin{split}   
        \mathcal{L}_{\text{triplet}}&=\sum_{i \in \text{Anchor}} \sigma\Big(\sum_{j \in \text{pos}} f_{a}({||h_{i}^{z} - h_{j}^{z^{'}}||}_{2})\\&- \sum_{j \in \text{neg}}f_{a}({||h_{i}^{z} - h_{j}^{z^{'}}||}_{2}) + \text{margin} \Big)
        \end{split}
\end{equation}

Anchor nodes are obtained by:
\begin{equation}
        \begin{split}   
        \text{Anchor}=TopK\left(f_{a}\left(A^{\text{sem}}\right) + f_{m}\left(A^{\text{sem}}\right)\right)
        \end{split}
\end{equation}
Here $z,z' \in \{dep, con\}$, and the anchor set size $k$ follows Bourgain’s Theorem-1 \cite{DBLP:conf/icml/YouYL19}, where $k=c\log^{2}n$. Functions $f_{a}$ and $f_{m}$ denote average and max pooling, “margin” controls triplet separation, and $\sigma$ is the ReLU activation.

\paragraph{Orthogonal Projection Techniques} We first project the dependency feature $H^{\text{dep}}$ onto the semantic feature $H^{\text{sem}}$:
\begin{equation}
        \begin{split}   
        H^{\text{dep}^{*}} = {Proj}\big(H^{\text{dep}}, H^{\text{sem}}\big)
        \end{split}
\end{equation}
with the projection function defined as:
\begin{equation}
        \begin{split}   
        {Proj}(x, y)=\frac{x \cdot y}{|y|} \frac{y}{|y|}
        \end{split}
\end{equation}
We then project $H^{\text{dep}}$ onto the orthogonal direction of $(H^{\text{dep}} - H^{\text{dep}^{*}})$ to obtain a cleaner representation:
\begin{equation}
        \widetilde{H^{\text{dep}}} = Proj\big(H^{\text{dep}}, (H^{\text{dep}} - H^{\text{dep}^{*}})\big)
\end{equation}
Similarly, the orthogonal projection for constituent features is:
\begin{equation}
        \widetilde{H^{\text{con}}} = Proj\big(H^{\text{con}}, (H^{\text{con}} - H^{\text{con}^{*}})\big)
\end{equation}

\subsubsection{Fusion Stage} 
In the fusion stage, we use the purified dependency feature $\widetilde{H^{\text{dep}}}$, purified constituent feature $\widetilde{H^{\text{con}}}$, semantic feature $H^{\text{sem}}$, and knowledge feature $H^{\text{kge}}$. Following MAMN \cite{DBLP:journals/tkde/XueZNW23, xue-etal-2023-constrained}, we apply its extended multimodal factorized bilinear pooling:
\begin{equation}
\begin{aligned}
        \mathcal{Z}_{m}^{i} &= Norm\left(\tilde{\mathrm{U}}_{\text{dep}}^{T} \widetilde{H^{\text{dep}}} \circ \tilde{\mathrm{U}}_{\text{con}}^{T} \widetilde{H^{\text{con}}} \right. \\
        & \left. \circ \tilde{\mathrm{U}}_{\text{sem}}^{T}{H}^{\text{sem}} \circ \tilde{\mathrm{U}}_{\text{kge}}^{T} {H}^{\text{kge}}\right)
\end{aligned}
\end{equation}
Here, $\tilde{U}$ denotes learnable projection matrices, $Norm$ is the normalization layer, and $\mathcal{Z}_{m}^{i}$ is the output of the $i$-th EMSF block. Residual connections are applied between blocks. The outputs from $l_{e}$ EMSF blocks are averaged:
\begin{equation}
\begin{aligned}
\mathcal{Z}_{m}^{i+1} & = \mathcal{Z}_{m}^{i} + \operatorname{EMSF}(\\ &\mathcal{Z}_{m}^{i}, \widetilde{H^{\text{dep}}}, \widetilde{H^{\text{con}}}, H^{\text{sem}}, H^{\text{kge}})
\end{aligned}
\end{equation}
The final EMGF output $r$ is obtained by averaging the outputs of the $l_{m}$ EMSF blocks:
\begin{equation}
r=\operatorname{Mean}\left(\mathcal{Z}_{m}^{1}, \mathcal{Z}_{m}^{2}, \ldots, \mathcal{Z}_{m}^{l_{m}}\right) 
\end{equation}

\subsubsection{Model Training}
\begin{table}[t]
\setlength{\tabcolsep}{2pt}
\small
\centering
\begin{center}
\renewcommand{\arraystretch}{1.3} 
\begin{tabular}{cccccccccc}
\hline
\multirow{2}{*}{\textbf{Dataset}} & \multicolumn{3}{c}{\#\textbf{Positve}} & \multicolumn{3}{c}{\#\textbf{Negative}} & \multicolumn{3}{c}{\#\textbf{Neutral}} \\
\cline { 2 - 10 }
 & Train & Dev & Test & Train & Dev & Test & Train & Dev & Test \\
\hline
Laptop & 976 & - & 337 & 851 & - & 128 & 455 & - & 167 \\
Restaurant & 2164 & - & 727 & 807 & - & 196 & 637 & - & 196 \\
Twitter & 1507 & - & 172 & 1528 & - & 169 & 3016 & - & 336 \\
MAMS & 3380 & 403 & 400 & 2764 & 325 & 329 & 5042 & 604 & 607 \\
\hline
\end{tabular}
\caption{Satistics of four datasets.}
\label{datasets}
\end{center}
\end{table}

\begin{table*}[h]
	\footnotesize
        \renewcommand{\arraystretch}{1.3}
	\centering
	\resizebox{\textwidth}{!}{\begin{tabular}{lcccccccc}
            \hline
            \multirow{2}{*}{\textbf{Model}} & \multicolumn{2}{c}{\textbf{Laptop}} & \multicolumn{2}{c}{\textbf{Restaurant}} & \multicolumn{2}{c}{\textbf{Twitter}} & \multicolumn{2}{c}{\textbf{MAMS}}\\
            \cline { 2 - 9 }
             & \textbf{Accuracy} & \textbf{Macro-F1} & \textbf{Accuracy} & \textbf{Macro-F1} & \textbf{Accuracy} & \textbf{Macro-F1} & \textbf{Accuracy} & \textbf{Macro-F1}\\
            \hline

            SSEGCN \cite{zhang-etal-2022-ssegcn} & 81.01 & 77.96 & 87.31 & 81.09 & 77.40 & 76.02 & - & -\\
            MGFN \cite{tang-etal-2022-affective} & 81.83 & 78.26 & 87.31 & 82.37 & 78.28 & 77.27 & - & -\\
            TF-BERT \cite{zhang-etal-2023-span} & 81.49 & 78.30 & 86.95 & 81.43 & 77.84 & 76.23 & - & -\\
            HyCxG \cite{xu-etal-2023-enhancing} & 82.29 & 79.11 & 87.32 & 82.24 & - & - & 85.03 & 84.40\\

            DAGCN \cite{DAGCN}
            & 78.96 & 75.07
            & 84.72 & 78.08
            & 77.10 & 75.66 
            & - & -\\
            MambaForGCN \cite{MambaForGCN}
            & 81.80 & 78.59
            & 86.68 & 80.86
            & 77.67 & 76.88 
            & - & -\\
            \hline

            CausalEdit-ABSC \cite{Li_Wang_Zhao_Zhang_Li_2025} 
            & 76.70 & -
            & 88.10 & -
            & - & - 
            & - & -\\
            InstructABSC \cite{InstructABSA} 
            & 81.56 & -
            & 86.25 & -
            & - & - 
            & - & -\\     
            LLaMA-CD-ABSA \cite{LLaMACD}
            & 83.70 & 81.02
            & 88.73 & 83.02
            & - & - 
            & $\textbf{86.74}$ & $\textbf{86.13}$\\
            LLaMA-3-8B-LoRA
            & 84.48 & 81.33
            & 90.62 &  85.10
            & - & - 
            & - & -\\   
            \hline
            \textbf{EMGF} 
            & 83.83 & 80.80
            & 88.42 & 83.20
            & 77.70 & 77.09 
            & 85.48 & 84.85\\
            \textbf{EMGF+CL} 
            & $\textbf{85.76}$ & $\textbf{82.98}$ 
            & $\textbf{90.80}$ & $\textbf{85.75}$ 
            & $\textbf{78.29}$ & $\textbf{77.45}$ 
            & 86.30 & 85.87\\

            \hline
            \end{tabular}}      
	\caption{Experimental results. For the Laptop, Restaurant, and MAMS datasets, we use Dependency Complexity as the curriculum learning difficulty indicator, while for the Twitter dataset, we adopt Sentence Length.}
	\label{main results}
\end{table*}

\paragraph{Softmax Classifier}
The fused representation $r$ is fed into a linear layer with a softmax function to obtain the sentiment distribution:
\begin{equation}
        \hat{y}_{(s, a)}=Softmax\left(W_{p}r+b_{p}\right)
\end{equation}
where $(s, a)$ is a sentence-aspect pair. The overall training objective combines the classification loss and the triplet loss:
\begin{equation}
        \mathcal{L}(\Theta)=\mathcal{L}_{c}+\beta \mathcal{L}_{\text{triplet}}
\end{equation}
where $\Theta$ denotes all trainable parameters and $\beta$ is a hyperparameter. The cross-entropy loss is:
\begin{equation}
        \mathcal{L}_{c}=\sum_{(s, a) \in \mathcal{D}} y_{(s, a)} \log \hat{y}_{(s, a)}
\end{equation}
where $\mathcal{D}$ is the set of all sentence–aspect pairs and $y_{(s,a)}$ is the ground-truth sentiment distribution.

\subsection{Curriculum Learning}
Curriculum learning \cite{CL} organizes training samples in an easy-to-hard manner, enabling the model to progressively learn complex semantic and syntactic patterns. In contrast to prior ABSA studies, which treat all samples formed by a sentence and its corresponding aspect as equally difficult, we are the first to propose a task-specific difficulty modeling framework for text-only ABSA.

\subsubsection{Difficulty Measure Function}
For each sample, we define a difficulty score based on five indicators:
\begin{itemize}
    \item \textbf{Sentence length} $d_{\text{len}}$: longer sentences tend to contain more clauses and potential opinion targets. We set
    $d_{\text{len}} = \min(\lvert s\rvert / 80,\, 1.0)$, where $\lvert s\rvert$ is the token length.
    \item \textbf{Aspect count} $d_{\text{asp}}$: sentences with more aspects are harder due to aspect interactions. We set
    $d_{\text{asp}} = \min((N_{\text{asp}} - 1)/4,\, 1.0)$, where $N_{\text{asp}}$ is the number of aspects in $s$.
    \item \textbf{Polarity pattern} $d_{\text{pol}}$: following the intuition that neutral and conflicting sentiments are harder, we assign
    $d_{\text{pol}} = 0.8$ if any aspect is neutral, $d_{\text{pol}} = 0.6$ if both positive and negative aspects appear, and $d_{\text{pol}} = 0.2$ otherwise.
    \item \textbf{Dependency complexity} $d_{\text{syn}}$: we measure the average and maximum dependency distance from the parsed tree, and map them into $[0,1]$:
    \[
        d_{\text{syn}} = \min\Big(\tfrac{1}{2}\big(\overline{d}/8 + d_{\max}/15\big),\, 1.0\Big),
    \]
    where $\overline{d}$ and $d_{\max}$ denote the mean and maximum absolute head--modifier distance.
    \item \textbf{Aspect position} $d_{\text{pos}}$: aspects appearing later in the sentence are usually harder to predict due to longer contextual dependencies. We compute the normalized average aspect start index as $d_{\text{pos}} \in [0,1]$.
\end{itemize}

We perform curriculum learning using each difficulty indicator individually and design a composite difficulty score with equal weights for all five indicators as a baseline. Their effectiveness is analyzed in the experiments.

\subsubsection{Training Scheduler}

Following standard curriculum-learning procedures, we first compute a difficulty score for each sample. The dataset $D$ is then sorted by difficulty and partitioned into $n$ buckets:
\[
    D_1, D_2, \ldots, D_n, \quad \text{with } \max(D_i) \le \min(D_{i+1}).
\]
Training begins with the easiest bucket $D_1$, and progressively incorporates more difficult subsets. After all buckets are introduced, several additional epochs are performed on the full dataset to refine the model.

\section{Experimental Settings} 

\begin{table*}[h]
	\footnotesize
        \renewcommand{\arraystretch}{1.4}
	\centering
	\resizebox{\textwidth}{!}{\begin{tabular}{lcccccccc}
            \hline
            \multirow{2}{*}{\textbf{Model}} & \multicolumn{2}{c}{\textbf{Laptop}} & \multicolumn{2}{c}{\textbf{Restaurant}} & \multicolumn{2}{c}{\textbf{Twitter}} & \multicolumn{2}{c}{\textbf{MAMS}}\\
            \cline { 2 - 9 }
             & \textbf{Accuracy} & \textbf{Macro-F1} & \textbf{Accuracy} & \textbf{Macro-F1} & \textbf{Accuracy} & \textbf{Macro-F1} & \textbf{Accuracy} & \textbf{Macro-F1}\\
            \hline
            \textbf{EMGF+CL}
            & $\textbf{85.76}$ & $\textbf{82.98}$ 
            & $\textbf{90.80}$ & $\textbf{85.75}$ 
            & $\textbf{78.29}$ & $\textbf{77.45}$ 
            & $\textbf{86.30}$ & $\textbf{85.87}$\\
            \hline
            EMGF-M4
            & 83.83 & 80.80 
            & 88.42 & 83.20 
            & 77.70 & 77.09 
            & 85.48 &84.85\\
            EMGF-M3 & 81.26 & 78.24 & 87.78 & 82.23 & 77.53 & 76.27 & 85.11 & 84.13\\
            EMGF-M2 & 80.79 & 77.61 & 87.33 & 81.59 & 76.64 & 76.12 & 84.32 & 83.75\\
            EMGF-M1 & 80.15 & 77.07 & 86.24 & 80.12 & 76.49 & 75.05 & 83.34 & 82.73\\
            \hline
            \emph{W/O} $\mathcal{L}_{\text{triplet}}$ & 80.84 & 76.83 & 86.97 & 81.06 & 76.93 & 75.61 & 83.54 & 83.21\\
            \hline
            \emph{W/O} Orthogonal Projection & 79.41 & 75.24 & 86.15 & 80.22 & 77.83 & 76.53 & 84.44 & 84.13\\
            \emph{W/O} Dep Project Sem & 80.52 & 76.92 & 86.15 & 79.96 & 76.04 & 75.20 & 84.44 & 83.87\\
            \emph{W/O} Con Project Sem & 80.37 & 76.47 & 85.70 & 79.66 & 76.19 & 74.98 & 83.99 & 83.48\\
            \hline
            \end{tabular}}      
	\caption{Ablation study experimental results. EMGF+CL is based on M4, with curriculum learning (CL)}
	\label{Ablation study}
\end{table*}

\subsection{Datasets}
We evaluate our model on four benchmark datasets: Laptop and Restaurant from SemEval2014 Task~4 \cite{pontiki-etal-2014-semeval}, Twitter \cite{dong-etal-2014-adaptive}, and the multi-aspect MAMS dataset \cite{jiang-etal-2019-challenge}. Following prior work \cite{chen-etal-2017-recurrent, li-etal-2021-dual-graph, tang-etal-2022-affective}, instances labeled as "conflict" are removed. Dataset statistics are provided in Table~\ref{datasets}.

\subsection{Implementation Details}
We use SuPar\footnote{\url{https://github.com/yzhangcs/parser}} to obtain dependency and constituent trees, and adopt the DeBERTaV3 \cite{he2023debertav} encoder with a dropout rate of 0.3. Training is performed with a batch size of 16, the Adam optimizer, and a learning rate of 2e\text{-}5. For the four datasets, ConGCN, DepGCN, and SemGCN layers are configured as (6, 3, 6, 6), (3, 3, 9, 3), and (3, 3, 1, 3), with $\beta$ values of (0.12, 0.12, 0.07, 0.12). We set $l_{c}=3$ constituent layers and $l_{e}=6$ EMSF blocks, and use a margin of~0.2. For curriculum learning, training samples are divided into 6, 8, 7, and 7 difficulty-based buckets for the four datasets. All experiments are repeated five times and averaged. Accuracy (Acc.) and macro-F1 (F1) are used as evaluation metrics.

\subsection{Baseline Methods}
We compare EMGF with representative baselines, grouped into small and large models.

The small models, including SSEGCN \cite{zhang-etal-2022-ssegcn}, MGFN \cite{tang-etal-2022-affective}, TF-BERT \cite{zhang-etal-2023-span}, HyCxG \cite{xu-etal-2023-enhancing}, DAGCN \cite{DAGCN} and MambaForGCN \cite{MambaForGCN}

The large models, including CausalEdit-ABSC \cite{Li_Wang_Zhao_Zhang_Li_2025}, InstructABSA \cite{InstructABSA}, LLaMA-CD-ABSA \cite{LLaMACD} and LLaMA-3-8B fine-tuned with LoRA.


\section{Results and Analysis} 
\subsection{Comparison with the State of the Art} 
As shown in Table~\ref{main results}, EMGF+CL achieves the best performance on most datasets and ranks second on MAMS. Overall, the results demonstrate that our approach not only surpasses existing small-model methods but also outperforms several large-model baselines. Notably, even with a small model combined with curriculum learning, our method exceeds the performance of the advanced large model fine-tuned with LoRA: LLaMA-3-8B.

\subsection{Ablation Study} 
We evaluate EMGF’s extensibility by varying the number of granularity features, as shown in Table~\ref{Ablation study}, with curriculum learning providing significant performance gains. M4 uses all four granularity features, while M3, M2, and M1 select three, two, and one feature(s) respectively and average the results. Performance decreases as fewer features are used, confirming EMGF’s extensibility and the cumulative benefits of its fusion mechanism. Removing $\mathcal{L}_{\text{triplet}}$ reduces performance, indicating that multi-anchor triplet learning effectively captures complementary syntactic information. “Dep Project Sem / Con Project Sem” refers to projecting dependency or constituent features onto semantic-orthogonal spaces; removing either the projection or the entire orthogonal projection technique lowers performance, showing that these projections help disentangle syntactic and semantic information and reduce redundant interference during the fusion stage.

\begin{table*}[t]
\footnotesize
\renewcommand{\arraystretch}{1.4}
\centering
\resizebox{\textwidth}{!}{
\begin{tabular}{lcccccccc}
\hline
\multirow{2}{*}{\textbf{Difficulty Method}} 
& \multicolumn{2}{c}{\textbf{Laptop}} 
& \multicolumn{2}{c}{\textbf{Restaurant}}
& \multicolumn{2}{c}{\textbf{Twitter}}
& \multicolumn{2}{c}{\textbf{MAMS}} \\
\cline{2-9}
& \textbf{Accuracy} & \textbf{Macro-F1}
& \textbf{Accuracy} & \textbf{Macro-F1}
& \textbf{Accuracy} & \textbf{Macro-F1}
& \textbf{Accuracy} & \textbf{Macro-F1} \\
\hline
No Curriculum 
& 83.83 & 80.80
& 88.42 & 83.20
& 77.70 & 77.09 
& 85.48 & 84.85 \\
\hline
Sentence Length (\(d_{\text{len}}\)) 
& 84.81 & 81.89
& 90.26 & 85.13
& \textbf{78.29} & \textbf{77.45}
& 85.70 & 85.37 \\
Aspect Count (\(d_{\text{asp}}\)) 
& 84.81 & 81.88
& 89.99 & \textbf{85.95}
& 77.99 & 77.27
& 85.63 & 85.25 \\
Polarity Pattern (\(d_{\text{pol}}\)) 
& 84.34 & 81.38
& 88.92 & 83.67
& 77.40 & 76.72
& 85.93 & 85.49 \\
Dependency Complexity (\(d_{\text{syn}}\)) 
& \textbf{85.76} & \textbf{82.98}
& \textbf{90.80} & 85.75
& 77.99 & 76.91
& \textbf{86.30} & \textbf{85.87} \\
Aspect Position (\(d_{\text{pos}}\)) 
& 85.44 & 82.71
& 89.99 & 85.04
& 78.14 & 77.19
& 85.70 & 85.25 \\
Composite 
& 85.28 & 82.48
& 89.28 & 84.34
& 77.68 & 76.99
& 85.55 & 85.05 \\
\hline
\end{tabular}}
\caption{Accuracy and Macro-F1 of different difficulty modeling methods for curriculum learning across four datasets. "Composite" denotes the equally weighted combination of the five difficulty indicators.}
\label{tab:curriculum_avg}
\end{table*}

\begin{figure*}[t]
    \centering
    \includegraphics[width=1.9\columnwidth]{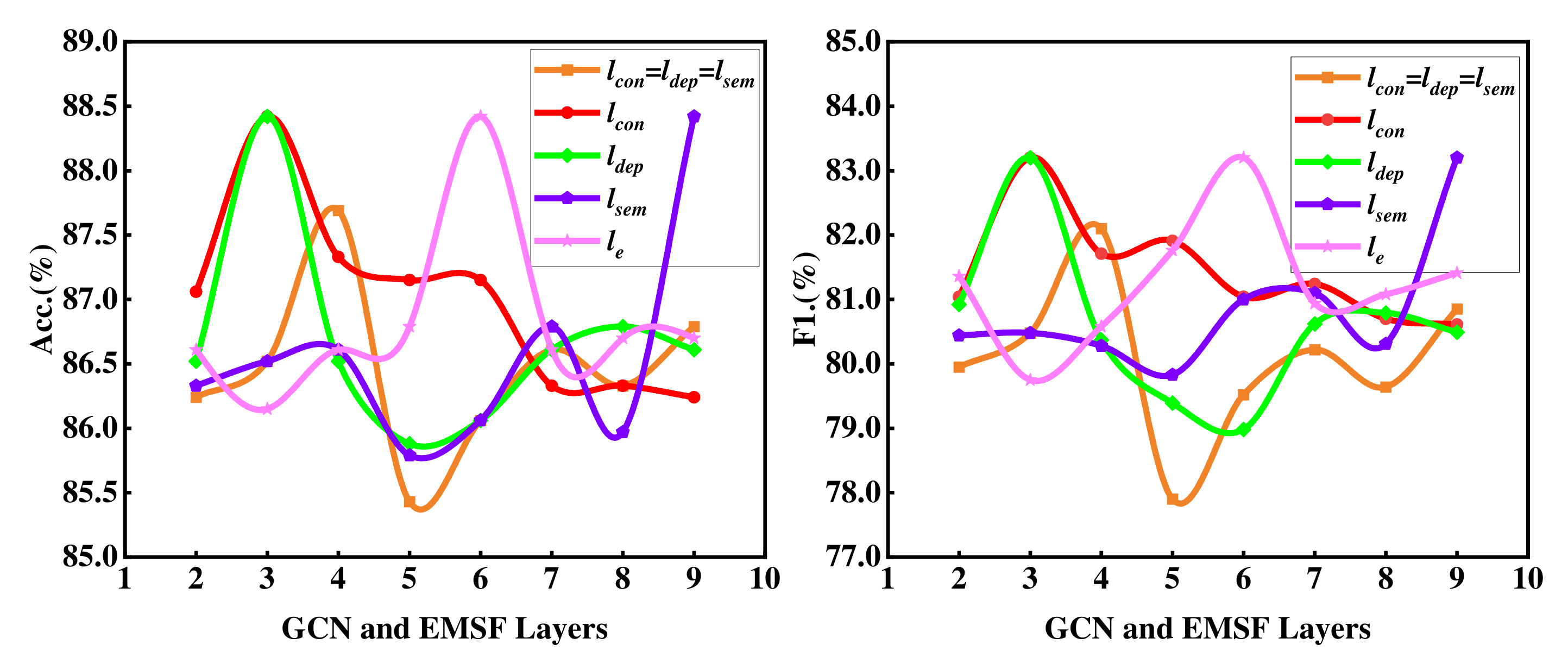}
    \caption{
    The impact of the number of GCN and EMSF block on Restaurant dataset.
    }
    \label{GCNLayer}
\end{figure*}
    
\subsection{Impact of Number of GCN and EMSF Blocks}
We vary $l_{con}$, $l_{dep}$, and $l_{sem}$ from 2 to 9 and find the optimal settings on the Restaurant dataset to be 3, 3, and 9. Using identical layer counts for all GCNs is suboptimal; tuning each independently yields better performance. The number of EMSF blocks $l_{e}$ also affects accuracy and F1, with the best setting being 6 (as shown in Figure~\ref{GCNLayer}).

\subsection{Hyper-parameter Analysis}
\begin{figure}[t]
    \centering
    \includegraphics[width=1.0\columnwidth]{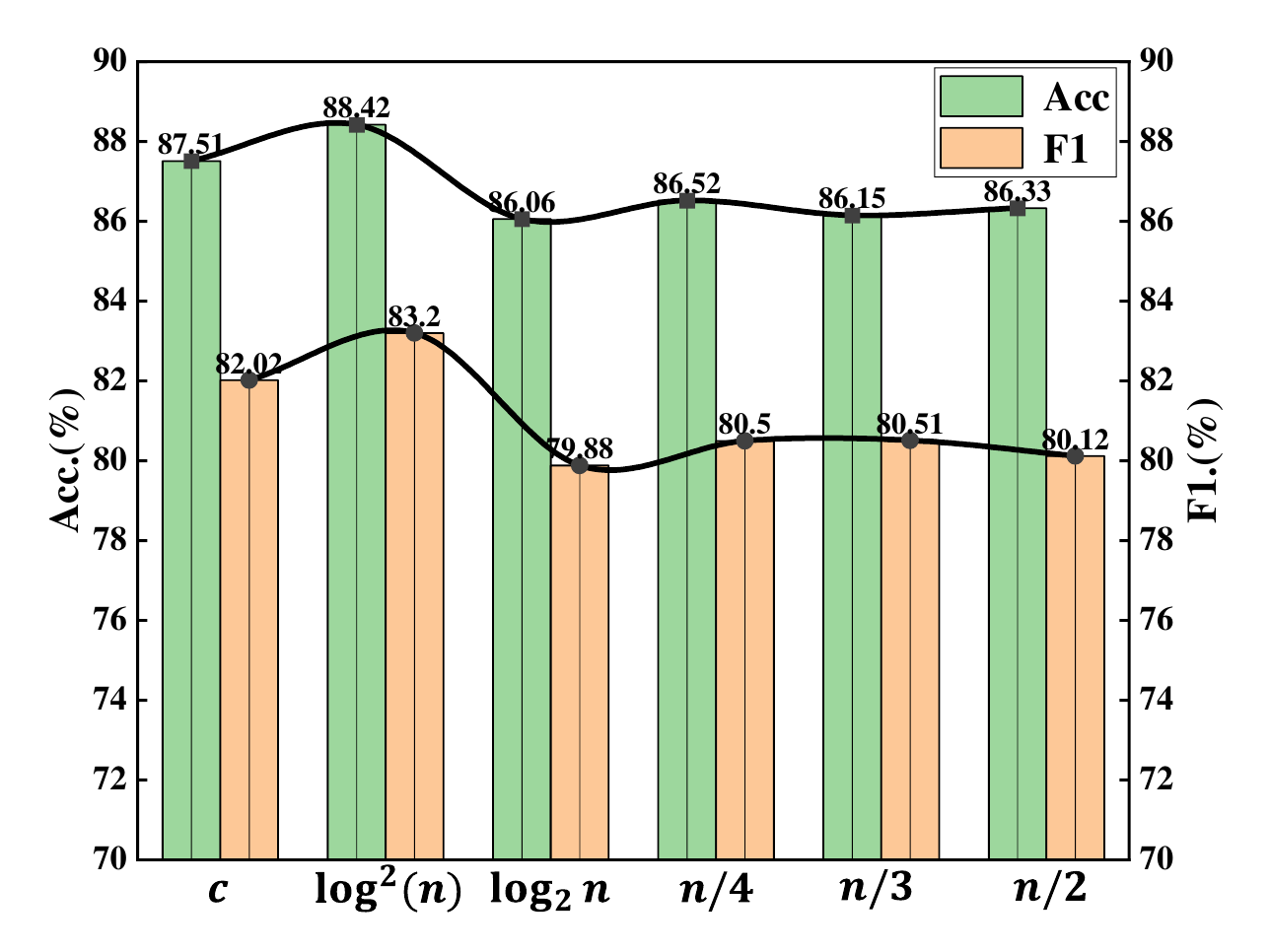}
    \caption{
    The impact of different $k$ on Restaurant dataset.
    }
    \label{k}
\end{figure}

We analyze the hyper-parameter $k$, which controls the number of selected anchor nodes in each view. Following Bourgain’s theorem~\cite{DBLP:conf/icml/YouYL19}, $k$ is set as $k=c\log^{2}(n)$, where $n$ is the number of nodes and $c$ (ranging from 1 to 5) adjusts the anchor scale. We further compare multiple settings of $k$, including $\{c, \log^{2}(n), \log_{2}(n), \frac{n}{4}, \frac{n}{3}, \frac{n}{2}\}$.  
As shown in Figure~\ref{k}, performance peaks when $k=\log^{2}(n)$, demonstrating its effectiveness.

\subsection{Analysis of Difficulty Modeling in Curriculum Learning}
As shown in Table~\ref{tab:curriculum_avg}, different difficulty indicators yield varying gains across datasets due to their distinct linguistic characteristics. For Laptop and Restaurant, which contain longer sentences and multiple aspects, syntactic-oriented indicators—especially \textit{dependency complexity}—provide the most significant improvements. In contrast, Twitter consists of short, noisy, and mostly single-aspect texts, making simple cues such as \textit{sentence length} more effective, while aspect-dependent indicators contribute little. MAMS, with its multi-aspect structure, also benefits notably from \textit{dependency complexity}. For comparison, we also employ a composite indicator obtained by the weighted average of the five difficulty metrics, but its overall performance is inferior to that of the best single indicator. Overall, these results show that curriculum learning is highly dataset-dependent, and selecting difficulty indicators that match the dataset’s characteristics is essential for achieving consistent performance gains. 

\section{Conclusion}
This study proposes EMGF+CL, addressing the long-standing challenge of fully leveraging the combined potential of multi-granularity features in the ABSA framework and innovatively introducing a curriculum learning strategy in the ABSA domain. EMGF effectively captures complex interactions among these features through multi-anchor triplet learning and orthogonal projection techniques, achieving cumulative benefits without additional computational cost. Moreover, by incorporating curriculum learning based on task-specific difficulty indicators, EMGF progressively guides the model from easier to more difficult samples, further enhancing learning efficiency and model performance.

\section*{Limitations}
Although our research has achieved commendable results, there are limitations worth acknowledging. These limitations underscore areas for future improvement and exploration. In this experiment, due to limited computational resources, we selected the top-$k$ nodes as Anchor nodes in multi-anchor triplet learning. However, when we attempted to set the value of $k$ to $\{\log_2{n}, \frac{n}{4}, \frac{n}{3}, \frac{n}{2}\}$ magnitude, we observed that the model training was excessively slow, and we had to adjust the magnitude of $k$ to a smaller scale for experimentation. Finally, due to constraints in computational power and time, we were unable to explore larger model architectures or conduct extensive hyperparameter tuning. 

Moreover, although curriculum learning significantly improves performance on most datasets, its effect on the Twitter dataset is not as pronounced due to the large variation in feature characteristics across ABSA datasets. Therefore, designing a more generalizable curriculum difficulty metric that can be consistently applied across all datasets remains a promising direction for future research.

\bibliography{custom}
\end{document}